\documentclass[journal, 10pt, twocolumn]{IEEEtran}

\usepackage{graphicx}
\usepackage[utf8]{inputenc}
\usepackage{booktabs} 
\usepackage{cite}
\usepackage{amsmath}
\usepackage{mathrsfs}
\usepackage{bm}
\usepackage[switch]{lineno} 
\usepackage{color}
\usepackage[T1]{fontenc}
\usepackage{comment}
\usepackage[justification=centering]{caption}
\usepackage{subfigure}
\usepackage{multirow}
\usepackage{amssymb}
\usepackage[numbers,sort&compress]{natbib}
\usepackage{bm}

\title{Towards Efficient Full 8-bit Integer DNN Online Training on Resource-limited Devices without Batch Normalization}

\author{Yukuan Yang$^{a}$, Xiaowei Chi$^{b}$, Lei Deng$^{a}$, Tianyi Yan$^{c}$, Feng Gao$^{d}$, Guoqi Li$^{a*}$

\small $^a$Department of Precision Instrument, Center for Brain Inspired Computing Research, Tsinghua University, Beijing 100084, China. \\ 
$^b$International School, Beijing University of Posts and Telecommunications, Beijing 100876, China.\\
$^c$School of Life Science, Beijing Institute of Technology, Beijing 100084, China.
\\$^d$Department of Interventional Neuroradiology, Beijing Tiantan Hospital, Capital Medical University, Beijing 100070, China.

\thanks{$^*$Corresponding author. Y. Yang and X. Chi contributed equally to this work. Email address: liguoqi@mail.tsinghua.edu.cn (G. Li).}
}

\begin{document}

\maketitle

\begin{abstract}
Huge computational costs brought by convolution and batch normalization (BN) have caused great challenges for the online training and corresponding applications of deep neural networks (DNNs), especially in resource-limited devices. Existing works only focus on the convolution or BN acceleration and no solution can alleviate both problems with satisfactory performance. Online training has gradually become a trend in resource-limited devices like mobile phones while there is still no complete technical scheme with acceptable model performance, processing speed, and computational cost. In this research, an efficient online-training quantization framework termed EOQ is proposed by combining Fixup initialization and a novel quantization scheme for DNN model compression and acceleration. Based on the proposed framework, we have successfully realized full 8-bit integer network training and removed BN in large-scale DNNs. Especially, weight updates are quantized to 8-bit integers for the first time. Theoretical analyses of EOQ utilizing Fixup initialization for removing BN have been further given using a novel Block Dynamical Isometry theory with weaker assumptions. Benefiting from rational quantization strategies and the absence of BN, the full 8-bit networks based on EOQ can achieve state-of-the-art accuracy and immense advantages in computational cost and processing speed. In addition to the huge advantages brought by quantization in convolution operations, 8-bit networks based on EOQ without BN can realize >70$\times$ lower in power, >18$\times$ faster in the processing speed compared with the traditional 32-bit floating-point BN inference process. What’s more, the design of deep learning chips can be profoundly simplified for the absence of unfriendly square root operations in BN. Beyond this, EOQ has been evidenced to be more advantageous in small-batch online training with fewer batch samples. In summary, the EOQ framework is specially designed for reducing the high cost of convolution and BN in network training, demonstrating a broad application prospect of online training in resource-limited devices.
\end{abstract}


\section{Introduction}

With the support of rapidly increasing computing capacity and data volume, deep neural networks (DNNs) \cite{alom2018history} have achieved remarkable successes in many tasks like image classification \cite{krizhevsky2012imagenet}, object detection \cite{ren2015faster,redmon2016you}, robotics \cite{pierson2017deep} and natural language processing \cite{devlin2018bert,brown2020language}, etc. Despite state-of-the-art results in such fields, there still exists one biggest problem in the application of DNNs on portable devices. In the most common application scenarios, the model must be trained on large computing clusters (GPUs \cite{li2016performance} or TPUs \cite{cass2019taking}) with super computing power and energy support and then re-implemented on portable devices with limited computational and energy resources. That is to say, once the trained model is downloaded to the portable devices, the portable devices can not re-train or fine-tune the model in the newly-collected data for better performance \cite{liu2019performance}. At the same time, the private data of users is difficult to be obtained and uploaded for the model training. In this case, online training comes into being for promoting the GPU or TPU pre-trained model performance. It works by further updating the model on the private datasets with more personal characteristics and the whole training process will only be executed on the user's own portable devices, solving the problem of private data acquisition. However, most portable devices with limited resources can't be provided with such online training ability for two reasons: the huge cost of computation and energy caused by large models; the complex operations during the forward and backward propagation of batch normalization (BN) \cite{ioffe2015batch}.

In pursuit of more diverse tasks and higher performance, the DNN model is getting deeper and larger for the reason that the model with more layers and parameters can better extract features from the input data and generate more accurate results \cite{he2016deep}. What's more, relatively large batch data is essential for model optimization. Meanwhile, along with the performance improvement brought by a larger model and batch size, the lower computing speed, the increasing computational resource, the huge energy consumption are limiting the wider application of DNNs severely. Most DNN models can only be trained and even inferred on cumbersome devices with supercomputing power and energy support. And the computing capacity of portable devices like smartphones and wearable devices is far from satisfying the memory and energy demand of such a huge calculation. 

At the same time, served as a default technique in most deep learning tasks, it has been widely demonstrated that BN is effective for DNN training with high performance and good robustness. Although studies on understanding why BN works are still underway \cite{bjorck2018understanding,santurkar2018does,luo2018towards}, the practical success of BN is indisputable. Whereas the number of operations and computational resources BN needs are much less than the convolution layer, the computation complexity of BN is much higher than that of convolution both forward and backward because BN requires square root operations. And these high-cost operations involving strong nonlinearity bring giant challenges for resource-limited devices such as the ASIC chips and FPGA. What's more, BN contains many costly reduction and element-wise operations which are hard to be executed in parallel. Last but not least, the backward propagation requires saving the pre-normalized activations, thus occupying roughly twice the memory as a non-BN network in the training phase \cite{bulo2018place}. For these reasons, BN becomes the most crucial part in non-convolution layers \cite{jung2018restructuring} because it involves about 58.5\% of the execution time and 90\% operations of non-convolution layers \cite{zhijie2020bactran}. Research also shows that BN lowers the overall training speed by $>47\%$ for deep ResNet and DenseNet models \cite{wu2018l1} and brings incredible difficulties in the special DNN inference and online training accelerator architecture design \cite{zhijie2020bactran}.

Many researchers have tried to address the challenging issues of the huge computational cost and the BN trap in the training process of DNN models. Among the existing model compressing methods trying to address the first issue such as compact model \cite{sandler2018mobilenetv2}, tensor decomposition \cite{novikov2015tensorizing,WU2020309,WANG2020215} and network sparsifying \cite{han2015learning}, etc., network quantization can decrease the computational cost and accelerate processing speed profoundly by converting floating-point values and floating-point multiply-accumulate (MAC) operations to fixed-point values and bit-wise operations \cite{courbariaux2016binarized,rastegari2016xnor}. Initial researches focus on the inference quantization for improving the inference processing speed of a well-trained model \cite{choukroun2019low,krishnamoorthi2018quantizing,wu2020integer,KULKARNI202128}. Such works quantize the inference data including weight, activation to low-bit integers for less memory consumption and faster computing speed. Recently, training quantization is becoming a much hotter field for the increasing interests in providing resource-limited devices with online training ability \cite{zhou2016dorefa,deng2018gxnor,wang2018training,micikevicius2017mixed,das2018mixed}. These works add the data quantization of the training process including error, gradient, and update, mainly aiming at decreasing the computational and energy cost in the backward propagation. Among these works, WAGEUBN \cite{yang2020training} have pushed the online training DNN model to a high level where all the data (weight, activation, gradient, error, update, and BN) involving the training the process has been quantized to 8-bit and the full 8-bit training framework has been successfully applied in the large-scale DNN model training. However, they only quantize BN and the operations that are complex and difficult to process in parallel still can't be avoided. 

As for the second issue of the BN trap, many researchers have also done lots of works to alleviate the problem. L1-BN \cite{wu2018l1} and QBP2 \cite{banner2018scalable} replace the standard deviation with L1-norm value and range of the data distribution, respectively. Even if the complex operations in calculating the standard deviation of mini-batch are avoided, the forward and backward propagation with high computational and memory cost is still existing. WAGE \cite{wu2018training} replace BN with a layer-wise scalar, however, it has been verified that it's hard to be implemented in large-scale DNN model training \cite{yang2020training}. As aforementioned, removing BN takes great advantages in computational resources, training speed, ASIC chip and FPGA compatibility, and online training accelerator design.
Recently, FixupNet \cite{zhang2019fixup} points out that the gradient norm of certain activations and weights in residual networks without normalization is lower bounded by some constant and this causes the gradient explosion. Based on this analysis, a novel initialization method is designed for updating the network with proper scale and depth independence to remove BN. However, the theoretical analysis of Fixup initialization is very coarse and only focuses on the prevention of gradient explosion. What's more, they have not considered the model compressing and accelerating with quantization. More importantly, small-batch online training is seldom involved in existing removing BN works.

As mentioned above, there is still a lack of methods in addressing both issues named huge computational cost and BN trap for the online training of resource-limited devices completely. 
In this work, we firstly propose a novel complete training quantization framework without BN by combing Fixup initialization and a novel quantization scheme to address both challenging issues named huge computational cost and BN trap for efficient online training. Using this framework, we have successfully removed BN, 
which can not only speed up the computation but also avoid the unfriendly square root operations for ASIC chips and FPGA. What's more, every data, both forward and backward in DNN training, including weights, activations, error, gradient, and update, is quantized to turn high-cost float-point operations into efficient fixed-point operations.
Then, to further validate the proposed work theoretically, we provide a comprehensive understanding of the scheme behind removing BN for the online training from the point of Block Dynamical Isometry \cite{chen2020comprehensive} with weaker assumptions.
Finally, extensive experiments have been done to verify the effectiveness of the proposed no BN quantization framework. Based on this framework, we have removed BN in large-scale DNNs and achieved a full 8-bit network with little performance degradation. More than this, we have also analyzed the hardware performance in FPGA. Experiment results show that with this complete quantization training framework, we can achieve about 4$\times$ memory saving, >3$\times$ and 9$\times$ faster in speed, 10$\times$ and 30$\times$ lower in power, 9$\times$ and 30$\times$ smaller in circuit area in the most executed multiplication and accumulation operations, respectively. More importantly, the proposed 8-bit framework has successfully removed BN, realizing >70$\times$ lower in power, >18$\times$ faster in processing speed compared with the traditional 32-bit floating-point BN and bringing immeasurable simplification to the design of deep learning chips.
The small-batch training experiment, which is much closer to the realistic scene, is also done to verify the performance and computational advantages of the online training framework without BN in practical applications. Our contributions can be threefold, which are summarized as follows:

\begin{itemize}

\item We address two main challenging issues named high computational cost and BN trap existing in current online training schemes via proposing a novel efficient online training quantization framework named EOQ without BN. Based on this framework combining Fixup initialization and a novel quantization scheme, the computational cost of BN and the unfriendly operations for ASIC chips and FPGA are removed, solving the BN trap perfectly and simplifying the deep learning chip design profoundly; every data, both forward and backward, is quantized and all float-point operations are transferred into fixed-point operations, reducing the computational resources for training and accelerating the processing speed greatly.  

\item Further than all previous works, we realize full 8-bit large-scale DNN training without BN. Especially, small batch online training is explored and weight updates are quantized to 8-bit integers for the first time with a novel quantization strategy.

\item We explain why BN works and how to remove BN theoretically with weaker assumptions from the point of Block Dynamical Isometry \cite{chen2020comprehensive}, making a deeper understanding of how Fixup initialization avoids gradient vanishing and explosion.

\item Extensive experiments have been done to validate the effectiveness of the proposed online training quantization framework without BN. Experiment results show that the full 8-bit DNNs without BN can achieve state-of-the-art accuracy with much fewer overheads in the ImageNet dataset compared with existing works. The full 8-bit quantization brings about 4$\times$ memory saving, >3$\times$ and 9$\times$ faster speed, 10$\times$ and 30$\times$ lower power, 9$\times$ and 30$\times$ smaller circuit area in the most convolution executed multiplication and accumulation operations and removing BN can realize >70$\times$ lower power, >18$\times$ faster processing speed compared with the traditional 32-bit floating-point BN. What's more, the small-batch training experiment is also done to further verify the performance of the proposed EOQ framework, demonstrating a broad application prospect in resource-limited devices. 
\end{itemize}

The organization of this paper is as follows: Section \ref{related works} introduces the related works of DNN quantization and BN acceleration; Section \ref{quantization framework} details the novel online-training quantization framework termed EOQ without BN and Section \ref{sec:theory provement} illustrates the theoretical analysis of Fixup initialization from the point of Block Dynamical Isometry \cite{chen2020comprehensive}; Section \ref{experiment} presents the experiment results of the proposed EOQ framework and the corresponding analyses; Section \ref{conclusion} summarizes this work and delivers the conclusion.

\section{Related Works}
\label{related works}

We aim to solve the two biggest problems named huge computational cost and BN trap that perplex the online training on resource-limited devices. DNN quantization and BN acceleration are two effective methods for alleviating these two problems and recent works are summarized as follows. However, there is still no complete solution combining them and making special optimization for resource-limited device online training.

\noindent \textbf{DNN Quantization:}
DNN quantization has been confirmed as an effective technique for shrinking model size and accelerating processing speed \cite{deng2020model}. Usually, the $\ge 8$-$bit$ fixed-point data quantization for inference will suffer negligible accuracy loss \cite{chen2014dadiannao,jouppi2017datacenter}. At the same time, more aggressive low-bit inference quantization can be seen to get an ultrahigh execution performance with a certain degree of accuracy degradation. Compared with inference quantization, training quantization is more challenging and attractive for researchers for its higher complexity and greater applicability. Three aspects domains the research of training quantization: quantization completeness, data precision, and model performance after quantization. Many works, such as DoReFa \cite{zhou2016dorefa}, GXNOR-Net \cite{deng2018gxnor}, QBP2 \cite{banner2018scalable}, etc., lack quantization completeness for the reason that they only quantize part of data, not all of it, in the training processing. Meanwhile, data precision signifies quantizing data to the lowest bit width for model compression and acceleration. However, recent works like MP \cite{micikevicius2017mixed}, MP-INT \cite{das2018mixed}, FP8 \cite{wang2018training} still rely on the 16-bit float-point, 16-bit fixed-point, and 8-bit float-point values, respectively. Finally, the accuracy of existing works, including WAGE \cite{wu2018training}, WAGEUBN \cite{yang2020training}, etc., usually cause a loss of accuracy greater than 5\% in large-scale networks and datasets.

\vspace{10pt}
\noindent \textbf{BN Acceleration:} Allowing a more casual initialization method, a faster convergence speed, and much deeper layers, BN has become one of the most favorite techniques in DNN training. However, considering the tremendous difficulties in processing speed and computational resources brought by BN, researchers start to focus on BN acceleration in the training process. The works of BN acceleration can be two-fold: reducing or avoiding the high-cost operations of BN and removing BN completely. The former realizes BN acceleration by reducing the number of BN operations \cite{chen2020comprehensive} or replacing the high-cost standard deviation operations with certain values like L1-norm value \cite{wu2018l1} and data range \cite{banner2018scalable}. 
The latter removes BN completely by proper initialization methods \cite{zhang2019fixup}, novel activation functions \cite{Klambauer2017_Advances}, small constants or learnable scalars \cite{hanin2018start,de2020batch,shao2020normalization}, weight Standardization \cite{Huang_2017_ICCV,brock2021characterizing}, and adaptive gradient clipping techniques \cite{brock2021high}, etc. However, all these techniques haven't been deeply integrated with other DNN acceleration technologies.

\section{Quantization Framework without BN}
\label{quantization framework}

The main idea of this research is to explore the full 8-bit integer DNNs without BN by combining Fixup initialization and novel quantization techniques for solving the BN trap and reducing the computational cost. Here we will first introduce the Fixup initialization method in Section \ref{sec:fixupinit} and then detail the full 8-bit online training framework without BN both forward and backward comprehensively in Section \ref{sec:quantization}.


\subsection{Fixup Initialization}
\label{sec:fixupinit}

For a long time, the huge computing burden brought by BN has plagued the development of deep learning. Many researches try to remove BN and Fixup initialization \cite{zhang2019fixup} has been confirmed as a simple and effective method. Researchers observe that the output variance of ResNet with standard initialization methods grows exponentially with depth and finally causes the failure of gradient exploding. One important role BN plays is to ensure that the output variance of each residual block does not change dramatically with depth and then it can naturally avoid the gradient exploding problem.

Base on this observation, Fixup initialization is more concerned about the scale of the update and the key conception of it is to design an initialization such that SGD updates to the network function are in the right scale and independent of the depth. For this purpose, they re-scale the weight layers inside the residual branches with a layer-wise coefficient related to the number of network residual blocks and the number of block layers to make the updates are in the right scale. The layer-wise coefficient $\alpha ^L _i$ is defined as 

\begin{equation}
    \alpha ^L _i = L^{-\frac{1}{2m_i-2}}
    \label{equ:fixupscale}
\end{equation}
where $L$ is the number of network residual blocks and $m_i$ is the layer number in the $i$-$th$ block. Besides, they add some scalar multipliers and bias in the residual block for better performance. Fixup initialization shows great advantages after removing BN. However, it has not been well integrated with DNN acceleration technologies and it is still a great challenge to utilize Fixup initialization without compression and acceleration when it comes to DNN online training on resource-limited devices.

\subsection{Efficient Online Training Quantization Framework}
\label{sec:quantization}

Inspired by Fixup initialization \cite{zhang2019fixup}, for the purpose of shrinking the model size and accelerating computing speed, here we'd like to introduce the efficient online-training quantization framework removing BN and we name it EOQ. Like most quantization studies, we follow the straight-through estimator (STE) method \cite{2013Estimating,courbariaux2016binarized,zhou2016dorefa} for the gradient propagation in the backward process. The EOQ framework only contains integer operations in the whole training process and makes the backward propagation as easy as the forward pass because of the absence of BN. To make it clear, we will first deliver the overview of the whole framework and then illustrate the quantization functions used in EOQ. Finally, we will detail the quantization strategies of different data in DNNs.

\vspace{10pt}
\subsubsection{Quantization Overview}
\ 

\begin{figure*}
    \centering
    \subfigure[]{\includegraphics[width=0.96\textwidth]{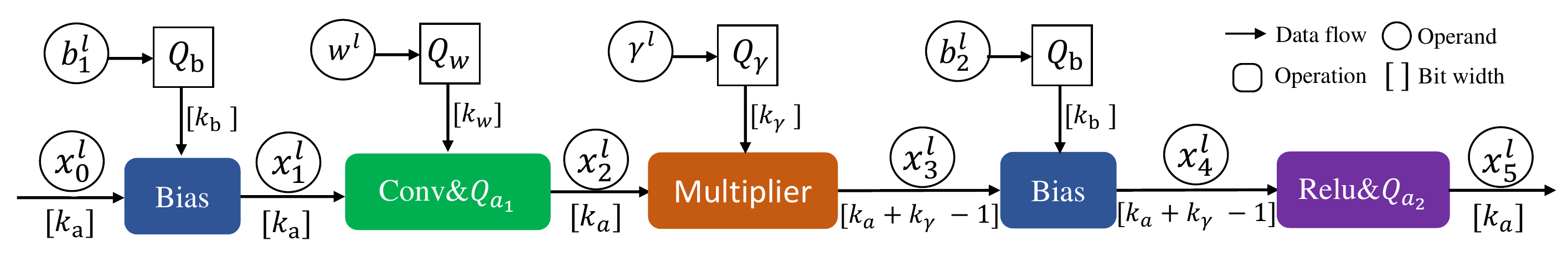} \label{fig:forward}}
    \subfigure[]{\includegraphics[width=0.96\textwidth]{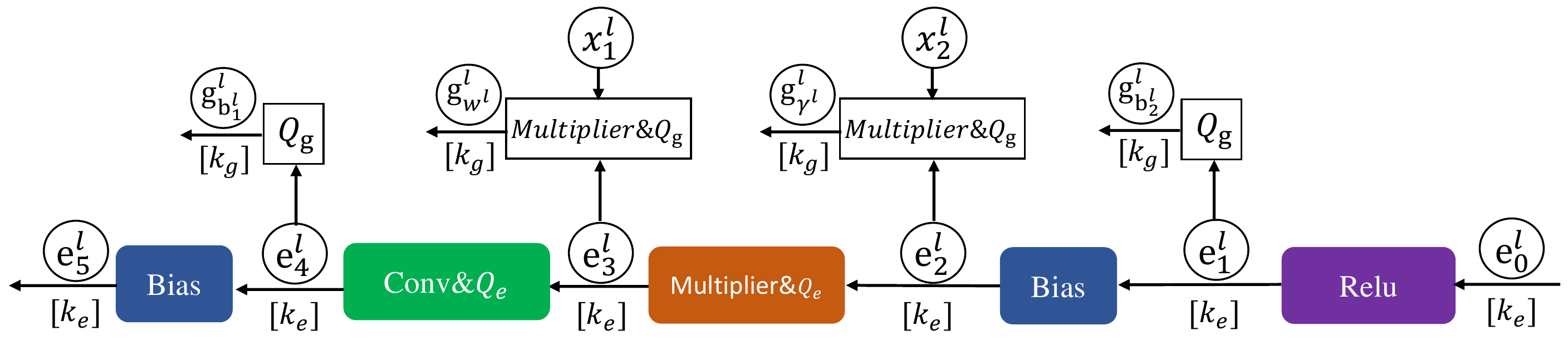} \label{fig:backward}}
    \caption{The overview of EOQ framework: (a) the forward propagation; (b) the backward propagation.}
    \label{fig:eoq}
\end{figure*}

The training process of DNNs can be divided into two parts: forward and backward propagation. 
Corresponding to the process and notations of Figure \ref{fig:forward}, for a typical Fixup residual block with bias and removing BN, the forward propagation can be represented as
\begin{equation}
\begin{split}
        &\bm{x}_0^l = \bm{x}_5^{l-1} \\
        &\bm{x}_1^l = \bm{x}_0^l + Q_b(\bm{b}_1)\\
        &\bm{x}_2^l = Q_{a_1}(conv(\bm{x}_1^l))\\
        &\bm{x}_3^l = Q_\gamma(\bm{\gamma}) \bm{x}_2^l\\
        &\bm{x}_4^l = \bm{x}_3^l + Q_b(\bm{b}_2)\\
        &\bm{x}_5^l = Q_{a_2}(relu(\bm{x}_4^l))
\end{split}
\end{equation}
where the superscript $l$ and the subscript $i\in \{1,2,3,4,5\}$ of $\bm{x}^l_i$ denote the layer index and the output of the $i$-$th$ operation block, respectively. $\bm{x}_5^{l-1}$ is the $(l-1)$-$th$ layer output and  $\bm{x}_0^l$ is the $l$-$th$ layer input. $\bm{b}_1$, $\bm{b}_2$ are bias and $\bm{\gamma}$ is a layer-wise scale. $Q_b$, $Q_\gamma$ and $Q_{a_{1}}$/$Q_{a_2}$ are the quantization functions for bias, layer-wise scale and activations, respectively, which will be detailed later in Section \ref{sec:quan_stra}. $k_w$, $k_b$, $k_\gamma$ and $k_a$ in Figure \ref{fig:forward} are the bit width of weights, bias, layer-wise scale and activations.

As shown in Figure \ref{fig:backward}, because of the absence of BN, the backward propagation is as simple as the forward propagation. The backward propagation can be summarized as
 \begin{equation}
      \begin{split}
      &\bm{e}^l_0=\bm{e}^{l+1}_5\\
      &\bm{e}^l_1=\bm{e}^l_0 \frac{\partial \bm{x}^5_l}{\partial \bm{x}^4_l} \\
      &\bm{e}^l_2=\bm{e}^l_1 \frac{\partial \bm{x}^4_l}{\partial \bm{x}^3_l} = \bm{e}^l_1 \\
      &\bm{e}^l_3=Q_e(\bm{e}^l_2 \frac{\partial \bm{x}^3_l}{\partial \bm{x}^2_l}) = Q_e(\bm{e}^l_2 Q_\gamma(\bm{\gamma}^l)) \\
      &\bm{e}^l_4=Q_e(\bm{e}^l_3 \frac{\partial \bm{x}^2_l}{\partial \bm{x}^1_l}) = Q_e(\bm{e}^l_3 Q_w(\bm{w}^l)) \\
      &\bm{e}^l_5=\bm{e}^l_4 \frac{\partial \bm{x}^1_l}{\partial \bm{x}^0_l} = \bm{e}^l_4 \\
      \end{split}
\end{equation}
where the superscript $l$ and the subscript $i\in \{1,2,3,4,5\}$ of $\bm{e}^l_i$ denote the layer index and the output of the last $i$-$th$ operation block, respectively. $\bm{e}_5^{l+1}$ is the $(l+1)$-$th$ layer error output and  $\bm{e}_0^l$ is the $l$-$th$ layer error input. $Q_e$ and $Q_w$ are the quantization functions for errors and weights that will be illustrated in Section \ref{sec:quan_stra}. $k_e$, $k_g$ in Figure \ref{fig:backward} are the bit width of errors and gradients.   
      
To compute and quantize the gradients of weight, bias and scale, we have
 \begin{equation}
      \begin{split}
      &\bm{g}^l_{w^l}=Q_g(\bm{e}^l_3 \frac{\partial \bm{x}_2^l}{\partial \bm{w}^l}) = Q_g(\bm{e}^l_3 \bm{x}^l_1) \\
      &\bm{g}^l_{b_1^l} = Q_g(\bm{e}^l_4 \frac{\partial \bm{x}^1_l}{\partial \bm{b}_1^l}) = Q_g(\bm{e}^l_4)  \\
      &\bm{g}^l_{b_2^l} =Q_g(\bm{e}^l_1  \frac{\partial \bm{x}^4_l}{\partial \bm{b}_2^l}) = Q_g(\bm{e}^l_1) \\
      &\bm{g}^l_{\gamma ^l}= Q_g(\bm{e}^l_2 \frac{\partial \bm{x}^3_l}{\partial \bm{\gamma} ^l}) = Q_g(\bm{e}^l_2 \bm{x}^l_2)  \\
      \end{split}
      \end{equation}
where $Q_g$ is the quantization function for gradients.

\vspace{10pt}

\subsubsection{Quantization Functions}
\label{sec:quant_func}
\ 

The quantization functions are used to convert the original floating-point values to fixed-point values. Considering both model performance and hardware compatibility, different data with diverse distributions in DNNs need to be quantized properly. Based on this observation, we propose three quantization functions for different data according to the corresponding distributions and effects on model performance.

\vspace{10pt}

\noindent \textbf{Basic quantization}: This basic quantization function $Q(.)$ is a fundamental part that approximates the floating-point value with its nearest integer. It works as 
\begin{equation}
    Q(\bm{x},k)=\frac{R(\bm{x} \cdot 2^{k-1})}{2^{k-1}}
\end{equation}
where $\bm{x}$, $R(\cdot)$ and $k$ are the quantized value, rounding function and quantized bit width, respectively. The resolution of basic quantization is $\frac{1}{2^k}$.

\vspace{10pt}

\noindent \textbf{Clamp quantization}: With the basic quantization function, we'd like to introduce the clamp quantization function to further limit the data range after quantization for hardware friendliness. Data after clamp quantization is constrained to $[-1+\dfrac{1}{2^{k-1}},1-\dfrac{1}{2^{k-1}}]$. The clamp quantization function is 
\begin{equation}
CQ(\bm{x},k)=Clamp(Q(\bm{x},k),-1+\frac{1}{2^{k-1}},1-\frac{1}{2^{k-1}})
\end{equation}
which has the same resolution as the basic quantization function.

\vspace{10pt}

\noindent  \textbf{Scale quantization}: The basic and clamp quantization is suitable for the data quantization with large values and low precision requirement. For the data  with small values and high precision demand, we propose the scale quantization as 
\begin{eqnarray}
\begin{split}
&~~~~~scale(\bm{x})=2^{R(\log_2(max(|\bm{x}|)))} \\
&SQ(\bm{x},k)=scale(\bm{x}) \cdot CQ(\frac{\bm{x}}{scale(\bm{x})},k)
\end{split}
\end{eqnarray}
whose resolution is $\frac{scale(\bm{x})}{2^k}$. The $scale(\cdot)$ function is used to estimate the maximum value of $\bm{x}$ and $CQ(\cdot)$ is used for normalization.  

\vspace{10pt}

 \subsubsection{Quantization Strategies} 
 \ 
 \label{sec:quan_stra}

Having given three quantization functions, here we will specify the quantization strategies for different data in DNNs. After analyzing the distributions of different data,
we find that the data in the forward propagation, such as weight, bias, activation, scale, etc., are with larger distribution range and lower precision requirements. While the data in the backward propagation, including error, gradient, and update, are with a smaller distribution range and higher precision requirements in the training process. Based on this, same as WAGEUBN \cite{yang2020training}, we use the basic and clamp quantization for the most data including weights, activations, scale, bias in the forward propagation, etc. and utilize scale quantization for error, gradient, and update quantization in the backward propagation empirically.

\vspace{10pt}

\noindent \textbf{Weight quantization:}
In the EOQ framework, the weight is initialized by Kaiming initialization \cite{he2015delving} in all layers and then scaled by $L^{-\frac{1}{2m_i-2}}$ only in the residual branch. To accelerate the computing and reduce the memory, the weight is quantized by
\begin{equation}
    Q_w(\bm{x}) = CQ(\bm{x},k_w)
\end{equation}
where $k_w$ is the bit width of the weight.

\vspace{10pt}

\noindent \textbf{Activation quantization:}
Activation is the most memory intensive among all data in DNNs. The bit width of activation is increased after convolution and multiplier. To further reduce the memory consumed by activation and accelerate computing, the activation is quantized by
\begin{eqnarray}
\centering
\begin{split}
    &Q_{a_1}(\bm{x}) = Q(\bm{x},k_a) \\
    &Q_{a_2}(\bm{x}) = CQ(\bm{x},k_a)
\end{split}
\end{eqnarray}
where $k_a$ is the bit width of activations. Here, to constrain the range of data transferred to the next layer, we use the clamp quantization after the layer activation function.

\vspace{10pt}

\noindent \textbf{Scale and bias quantization:} The scale and bias are used to replace traditional BN and further improve the model performance. They are initialized as 1 and 0 from the beginning. To realize a complete online training quantization framework and avoid the floating-point operations for hardware friendliness, we also quantize the scale and bias by
\begin{eqnarray}
\centering
\begin{split}
    &Q_{\gamma}(\bm{x}) = Q(\bm{x},k_\gamma) \\
    &Q_{b}(\bm{x}) = Q(\bm{x},k_b)
\end{split}
\end{eqnarray}
where $k_\gamma$ and $k_b$ are the bit width of the scale and bias, respectively.

\vspace{10pt}

\noindent \textbf{Error quantization:}
Because of the relatively high precision requirements, the quantization of error is most difficult and may make a great impact on the final model performance. Although WAGEUBN \cite{yang2020training} has solved the convergence problem when the bit width of error comes to 8, it still suffers a huge performance degradation (about 4\% and 7\% accuracy decline in ResNet-18 and ResNet-50 compared with the vanilla networks). After removing BN, the error can be quantized to 8-bit integers using the EOQ framework with nearly no accuracy loss in ResNet-18 and much less accuracy loss in ResNet-50. The error is quantized by
\begin{equation}
    Q_e(\bm{x}) = SQ(\bm{x},k_e)
\end{equation}
where $k_e$ is the bit width of error.

\vspace{10pt}

\noindent \textbf{Gradient quantization:}
The distribution of gradient is sometimes similar to that of error. The error needs to be backward propagated accurately layer by layer for loss descent while gradients are more important to maintain the magnitude and sign because it is used for weight update in a training step \cite{yang2020training}. The quantization of gradients is governed by
\begin{equation}
    Q_g(\bm{x}) = SQ(\bm{x},k_g)
\end{equation}
where $k_g$ is the bit width of gradients.

\vspace{10pt}

\noindent \textbf{Update quantization:}
Updating weight, scale, and bias is the final procedure in a training step.  Different from WAGEUBN \cite{yang2020training} quantizing updates to 24-bit, we first realize the 8-bit update quantization in the proposed EOQ framework. That is to say, the weight can be stored with 8-bit integers initially with a scale, achieving roughly $2\times$ weight memory space-saving. The update is quantized by
\begin{equation}
    Q_u(\bm{x})=SQ(\bm{x},k_u)=SQ(lr \cdot \bm{g},k_u)
\end{equation}
where $k_u$ is the bit width of the update, $lr$ is the learning rate decreased gradually during training, and $\bm{g}$ is the gradient of the updated data.

\section{Novel Theory Analysis of Removing BN}
\label{sec:theory provement}

Although the theoretical understanding of BN is still not unified \cite{bjorck2018understanding,santurkar2018does,luo2018towards}, we'd like to give a deeper understanding of why BN works in DNNs and how to remove BN in the EOQ framework using the novel Block Dynamical Isometry theory \cite{chen2020comprehensive} with weaker assumptions and a more intuitive point of view. At the same time, EOQ tries to estimate the original floating-point values with sufficiently precise fixed-point values. Therefore, in this research, the error caused by quantization can be neglected in the analysis of the quantization network with batch normalization and Fixup initialization. In this section, we will first introduce the indicator named gradient norm for judging whether gradients explode or vanish in DNNs. Next, the mathematical prerequisites are presented. Finally, we will provide the detailed training stability analyses of networks with batch normalization and Fixup initialization respectively using the Block Dynamical Isometry theory \cite{chen2020comprehensive}.

\vspace{10pt}
\subsubsection{Gradient Norm}
\label{gradirnt_norm}
\

Regarding each block as a function, considering a network with $L$ sequential blocks, we have

\begin{equation}
\label{seq_block}
f(x_0)=f_L(f_{L-1}(f_{L-2}(...f_1(x_0))))
\end{equation}
 where $f_i$ denotes the $i$-$th$ block, $x_0$ is the input of the network.

In the back propagation process, the update of weights in the $i$-$th$ block is
\begin{equation}
\Delta\theta_i=\eta\left(\frac{d f_{i}}{d \theta_{i}}\right)^{T} \left(\prod_{j=L}^{i+1} \left(\frac{d f_{j}}{d f_{j-1}}\right)\right)^{T}\frac{d \mathcal{L}(f_L(x),y)}{d f_L(x)}
\end{equation}
where $\mathcal{L}(\cdot)$ is the loss function; $\theta_i$ is the weights of the $i$-$th$ block $f_i$; $\eta$ is the learning rate.

Denoting the Jacobin matrix $\dfrac{d f_{j+1}}{d f_{j}}$ as $J_j$ and $\dfrac{d\mathcal{L} (f_L(x),y)}{d f_{L}(x)}$ as $J_L$ , the L2 norm of $\Delta\theta_i$ can be represented as

\begin{small}
\begin{equation}
\label{gradnorm}
\begin{split}
&||\Delta\theta_i||_2^2= \eta^2\left(J_L\right)^T\left(\prod_{j=L}^{i+1} J_i\right)\frac{d f_{i}}{d \theta_{i}}\left(\frac{d f_{i}}{d \theta_{i}}\right)^{T} \left(\prod_{j=L}^{i+1} J_i\right)^{T}J_L
\end{split}
\end{equation}
\end{small}

For a network with $E[\left||\Delta\theta_i\right||_2^2]$ (the expectation of $\left||\Delta\theta_i\right||_2^2$) tending to be $0$ or $\infty$, the gradient vanishing or explosion will happen and cause the failure of network training. Suppose $\eta$ and $J_L$ are constants, it's essential for network to be stable for training that the middle part $\left(\prod_{j=L}^{i+1} J_i\right)\dfrac{d f_{i}}{d \theta_{i}}\left(\dfrac{d f_{i}}{d \theta_{i}}\right)^{T} \left(\prod_{j=L}^{i+1}J_i\right)^T$ in Equation (\ref{gradnorm}) is not exponential related to the the block index $i$. 

\vspace{10pt}
\subsubsection{Prerequisites}
\ 

Before giving the mathematical analysis of BN and Fixup initialization in residual networks, we'd like to introduce some necessary definitions and lemmas.

\textbf{Definition 1. Normalized trace of a matrix:} In linear algebra, the sum of the elements on the main diagonal of an $n\times n$ matrix $A$ is called the trace of matrix $A$ and is generally denoted as $Tr(A)$. Here we denote the normalized trace of $A$ $\frac{Tr(A)}{n}$ as $tr(A)$, for example: $tr(I)=1$, where $I$ represents the identity matrix.

\textbf{Definition 2. Block Dynamical Isometry:} For a network represented as a sequence of individual blocks as Equation (\ref{seq_block}) and $J_j$ denoted as the $j$-$th$ block's Jacobian matrix, for all $j$, if $E\left[tr(J_j J_j^{T})\right]\approx 1$ and $D\left[tr(J_j J_j^{T})\right]\approx 0$, we say it achieves the Block Dynamical Isomtry \cite{chen2020comprehensive}, where $E[\cdot]$ and $D[\cdot]$ are the expectation and variance, respectively. Actually, in most cases, $E\left[tr(J_j J_j^{T})\right]\approx 1$ is enough to conclude the individual block satisfies the Block Dynamical Isomtry. If each block in a network satisfy Block Dynamical Isometry, the expectation of $\left||\Delta\theta_i\right||_2^2$ in Equation \ref{gradnorm} tending to be $0$ or $\infty$ is avoided and thus gradient vanishing or explosion will also disappear.

\textbf{Lemma 1. Variance of general transforms:} Given a general linear transform $f(x) = Jx$, then, we have $Var[f(x)]= E[tr(JJ^T)]Var[x]$, where $Var[\cdot]$ is the variance of $x$.

\textbf{Lemma 2. Additive transformation:} Given $J=\sum_{i}{J_i}$ where $J_i$  is independent random matrix. For all $J_i$, if at most one $E[J_i]\neq 0$, we have $E[tr(\sum_{i=L}^{1}{(J_iJ_i^T)})]=\sum_{i=L}^{1}{E[tr(J_iJ_i^T)]}$.

In summary, Definition 1 gives the fundamental mathematical indicator named normalized trace of a matrix, and Definition 2 illustrates the tool named Block Dynamical Isometry for judging whether gradient vanishing or explosion happens in network training. Lemma 1 and 2 are used in the analysis of BN and Fixup initialization and the proof of them can be found in the work \cite{chen2020comprehensive}.


\vspace{10pt}
\subsubsection{Training Stability Analysis of Network with Batch Normalization}
\

As mentioned in Section \ref{gradirnt_norm}, given a common network containing $L$ sequential blocks, we divide each block into three components, which are Relu activation, convolution, and BN, respectively.

\begin{figure}[!htbp]
    \centering
    \includegraphics[width=0.48\textwidth]{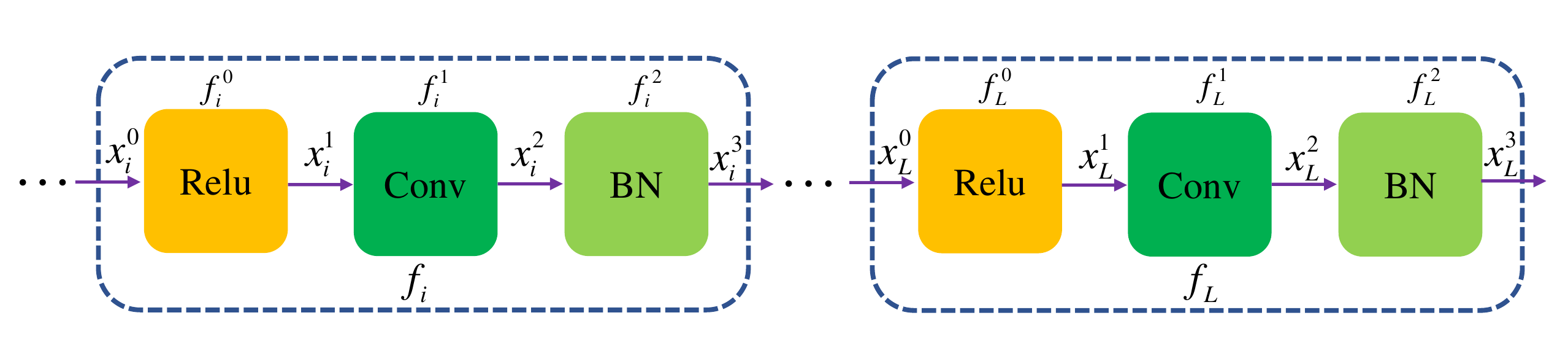}
    \caption{Networks composed of $L$ sequential blocks.}
    \label{fig:layer_block}
\end{figure}

As illustrated in Figure \ref{fig:layer_block}, we denote the Relu activation, convolution and BN in the $i$-$th$ block as $f_i^0$,  $f_i^1$,  $f_i^2$. Then we have

\begin{equation}
J_i^k = \frac{\partial f_i^k(x_i^k)}{\partial x_i^k}
\end{equation}

\noindent where $k$ $\in$ $\{0,1,2\}$.

Since Relu activation, convolution and BN can be seen as general linear transforms, then we have 

\begin{equation}
f_i^k(x_i^k)=J_i^k x_i^k.
\end{equation}

According to Lemma 1, supposing the variance of $x_i^k$ is $\alpha_i^k$, then, we have

\begin{equation}
\alpha^3_i= Var[f^3_i f^2_i( f^1_i(x_i^0))] = E[tr(  (\prod \limits_{k=2}^0J_i^k ) (  \prod \limits_{k=2}^0 J_i^k)^T)] \alpha_i^0
\end{equation}

Considering the $i$-$th$ block and denoting $\prod \limits_{k=2}^0 J_i^k$ as the Jacobian matrix of the block $J_i$, we have 
\begin{equation}
\alpha^3_i = \ E[tr(J_i (J_i)^T)] \alpha_i^0
\label{block_v}
\end{equation}

At the same time, in the initialization stage where $\gamma$ and $\beta$ in BN are initialized as 1 and 0, respectively, the variance of the BN output is 
\begin{equation}
\alpha^3_i = Var[f_i^3(x_i^2)] = 1
\label{const_v}
\end{equation}

Combing Equation \ref{block_v} and \ref{const_v}, we have
\begin{equation}
E[tr(J_i (J_i)^T)] = \dfrac{1}{\alpha_i^0}.
\end{equation}

Since $\alpha_i^0$ is also the variance of the output of BN, so $\alpha_i^0$ is 1 and wo get 
\begin{equation}
E[tr(J_i (J_i)^T)] = 1.
\end{equation}

Now we can conclude that the block illustrated as Figure \ref{fig:layer_block} containing BN satisfies Definition 2 and avoiding the vanishing and explosion of gradients.

\vspace{10pt}

\subsubsection{Training Stability Analysis of Network with Fixup Initialization}
\

After proving that BN satisfies the Block Dynamical Isometry theory, here we'd like to explain that Fixup initialization can also make a similar effect to BN and meet up this theory through re-scaling residual branches. The $i$-$th$ individual residual block in ResNet can be represented as a sum of the main branch $h_i(.)$ and residual branch $r_i(.)$, which can be represented as 

\begin{equation}
x_{i}=f_i(x_{i-1})=h_i(x_{i-1})+r_i(x_{i-1})
\end{equation}
where $i$ is the block index and $f_i$ denotes the $i$-$th$ block.

Considering the hybrid network block in series and parallel, according to Lemma 2, we have:

\begin{equation}
E(tr(J_i J_i^{T}))= E(tr(J_i^h (J_i^h)^T)) + E(tr(J_i^r (J_i^r)^T))
\end{equation}
Where $J_i^h$ and  $J_i^r$ are the Jacobian matrices of the main branch and residual branch, respectively. In FixupNet \cite{zhang2019fixup}, the weights $w$ in all layers are initialized with standard methods like Kaiming initialization \cite{he2015delving} to make the variance of the input keep consistent with that of the layer output. As illustrated in Equation \ref{equ:fixupscale}, then the weights inside residual branches are scaled with a coefficient $L^{-\frac{1}{2m_i-2}}$, where $L$ and $m_i$ is the number of residual blocks and layers inside the $i$-$th$ block, respectively. As one of the most popular initialization techniques, Kaiming initialization initializes weights to $N(0,\dfrac{2}{n})$ where $n$ is the number of layer input activations. So for the main branch  of the $l$-$th$ block using Kaiming initialization, we have 

\begin{equation}
E[tr(J^h_i (J_i^h)^T)]=1.
\label{equ:main_branch}
\end{equation}

For the $k$-$th$ layer in the residual branch of $i$-$th$ block using Kaiming initialization with a coefficient $L^{-\frac{1}{2m_i-2}}$, we have 
\begin{equation}
E[tr(J^h_{ik} (J_{ik}^{h})^T)]=L^{-\frac{1}{m_i-1}}.
\label{equ:res_branch_layer}
\end{equation}
where $m_i$ is usually a small positive number, e.g., 2 or 3.

And for the whole residual branch, we have 

\begin{eqnarray}
\centering
\begin{split}
E[tr(J^h_{i} (J_{i}^h)^T)] &= \prod \limits_{k=1}^{m_i} E[tr(J^h_{ik} (J_{ik}^h)^T) \\
&= L^{-\frac{m_i}{m_i-1}}
\end{split}
\end{eqnarray}

Next we can denote the $i$-$th$ block as
\begin{equation}
E(tr(J_i J_i^{T}))= 1+L^{-\frac{m_i}{m_i-1}}.
\end{equation}

Since $L$ is a relatively large positive integer especially in extremely deep residual networks, we can conclude 
\begin{equation}
E(tr(J_i J_i^{T}))= 1+L^{-\frac{m_i}{m_i-1}} \approx 1.
\end{equation}

Now we can testify that Fixup initialization helps to satisfy Definition 2 and is effective for training extremely deep residual neural networks without gradient vanishing and explosion.

\section{Experiments}
\label{experiment}

To verify the effectiveness of the proposed EOQ framework and build new full 8-bit DNNs removing BN, we have implemented extensive experiments and analyzed the performance impact and computational advantages caused by data quantization and removing BN. Experiment results show that the EOQ framework puts forward a complete solution for solving the two biggest challenging problems named huge computational cost and BN trap perfectly in DNN training.

To the best of our knowledge, We are the first to achieve the full 8-bit integer large-scale DNNs training without BN in the ImageNet dataset using the EOQ framework by setting $k_w=k_a=k_b=k_\gamma=k_e=k_g=k_u=8$ with much less accuracy loss. 
Results on the ImageNet dataset are given in Section \ref{sec:acc} and the computational advantages of EOQ are detailed in Section \ref{sec:advan}. 
Finally, Section \ref{sec:small} illustrates the results of small batch size experiments to show the superiority of EOQ in model performance. 

\subsection{Full 8-bit Training on the ImageNet Dataset}
\label{sec:acc}

We test the proposed EOQ framework with the ResNet-18/34/50 networks on the classification task over the ImageNet dataset. We go further than most works leaving the first and last layer unquantized \cite{zhou2016dorefa,cai2017deep,choi2018learning,yang2020training} by quantizing all layers in the model except the last layer to avoid the relatively huge accuracy loss. The comparisons between the full 8-bit integer EOQ networks and classical networks are as shown in Table \ref{tab:imagenet_acc}. To further improve model performance, we take advantage of the common data augmentation method $mixup$ and we set the hyperparameter $\alpha$ as $0.7$. The model is trained with synchronized stochastic gradient descent (SGD) optimizer over 4 GPUs and the total epoch number is 90.

\begin{figure}[htbp]
    \centering
    \includegraphics[width=0.5\textwidth]{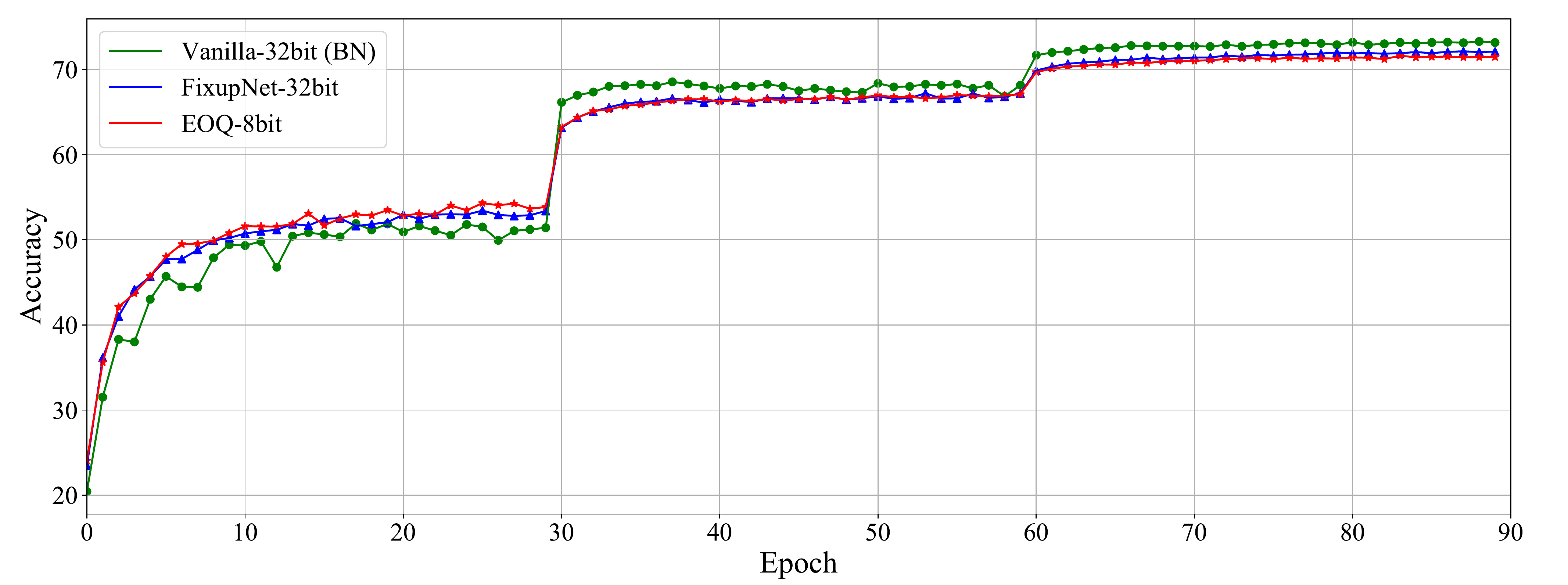}
    \caption{Validation accuracy of ResNet-34 during the training process.}
    \label{fig:resnet34_curve}
\end{figure}

Figure \ref{fig:resnet34_curve} shows the accuracy curves on validation set of ResNet-34 during the training process. We can see that there is almost no difference between the full 8-bit EOQ network and the 32-bit floating-point FixupNet. This also proves the rationality of quantization functions used in EOQ, indicating that the 8-bit integer data after quantization can even achieve the same performance and robustness as 32-bit float-point data in the whole training process. The results of EOQ and recent classical works on the ImageNet dataset are as shown in Table \ref{tab:imagenet_acc}. Compared with the floating-point vanilla network, the full 8-bit EOQ networks without BN only suffer 1.34\%, 1.92\%, and 2.99\% accuracy degradation in ResNet-18, 34, and 50 networks. While the corresponding data of full 8-bit WAGEUBN \cite{yang2020training} with BN are 3.91\%, 4.36\%, 6.71\%. Compared with the floating-point FixupNet \cite{zhang2019fixup} with data augmentation $mixup$, the full 8-bit EOQ networks take lower accuracy degradation, which are 0.56\%, 0.63\%, 3.28\% in ResNet-18, 34 and 50, respectively. Considering the huge improvements in computational cost brought by 8-bit quantization and removing BN compared with vanilla floating-point networks and the much less accuracy degradation compared with WAGEUBN \cite{yang2020training}, the EOQ framework has achieved a complete 8-bit training quantization scheme with high performance and excellent hardware efficiency.

\begin{table*}[!htpb]
    \centering
    \begin{tabular}{ccccc}
\hline  
Model &Method &BN &$k_w$/$k_a$/$k_g$/$k_e$/$k_u$ &Accuracy Top-1/5(\%) \\
\hline 
\multirow{6}*{ResNet-18}
&Vanilla   &\checkmark &32/32/32/32/32 & 68.70/88.37 \\
 &WAGEUBN &\checkmark &8/8/8/8/24 &64.79/84.80  \\
 &EOQ(ours) &$\times$ &8/8/8/8/8 &\textbf{67.36}/\textbf{87.72} \\
 &mixup &\checkmark &32/32/32/32/32 & 67.51/88.08\\
  &Fixup+mixup  &$\times$ &32/32/32/32/32 & 68.13/87.77  \\
&EOQ+mixup(ours) &$\times$ &8/8/8/8/8 &\textbf{67.57}/\textbf{87.62} \\
\hline
\multirow{6}*{ResNet-34}
&Vanilla   &\checkmark &32/32/32/32/32 & 71.99/90.56 \\
&WAGEUBN &\checkmark &8/8/8/8/24 &67.63/87.70 \\
&EOQ(ours) &$\times$ &8/8/8/8/8 &\textbf{70.07}/\textbf{89.22} \\
&mixup &\checkmark &32/32/32/32/32 & 73.17/91.34 \\
 &Fixup+mixup  &$\times$ &32/32/32/32/32 & 72.11/90.48 \\
&EOQ+mixup(ours) &$\times$ &8/8/8/8/8 &\textbf{71.48}/\textbf{90.29}  \\
\hline 
\multirow{6}*{ResNet-50}
&Vanilla   &\checkmark &32/32/32/32/32 & 74.66/92.13\\
&WAGEUBN &\checkmark &8/8/8/8/24 &67.95/88.01 \\
&EOQ(ours) &$\times$ &8/8/8/8/8 &\textbf{71.67}/\textbf{90.08}{}  \\
 &mixup &\checkmark &32/32/32/32/32 & 75.72/92.74\\
  &Fixup+mixup  &$\times$ &32/32/32/32/32 & 75.54/92.52  \\
&EOQ+mixup(ours) &$\times$ &8/8/8/8/8 &\textbf{72.26}/\textbf{90.78}  \\
\bottomrule 
\end{tabular}
    \caption{ImageNet test results of the EOQ framework.}
    \label{tab:imagenet_acc}
\end{table*}

\begin{table*}[!htpb]
\centering
\resizebox{2\columnwidth}{!}{
\begin{tabular}{c|c|c|c|c|c|c|c|c|c}

\hline
\multirow{2}*{Model} &\multirow{2}*{Method}  & \multicolumn{8}{c}{Memory (MB)} \\
\cline{3-10}
& &Batch size=1 & Batch size=2 & Batch size=4 & Batch size=8 &  Batch size=16 &  Batch size=32 &  Batch size=64 & Batch size=128 \\
\hline
\multirow{3}*{ResNet-18} & Vanilla-32bit &127.30 &210.12 &375.64 & 706.68 &1368.78 &2692.96 & 5341.33 &10638.07\\

& Vanilla-8bit &31.83 &52.53 &93.91 &176.67 &342.20 &673.24 &1335.33 &2559.52\\

&EOQ-8bit &25.64 &40.15 &69.15 &127.16 &243.19 &475.23 &939.32 &1867.51\\
\hline

\multirow{3}*{ResNet-50} &Vanalla-32bit & 472.52 &847.54 &1597.59 &3097.68 &6097.86 &12098.24 & 24098.98 &48100.47 \\

&Vanalla-8bit &118.13 &211.89 &399.40 &774.42 &1524.47 &3024.56 &6024.75 &12025.12 \\

&EOQ-8bit &90.39 &156.46 &288.60 &552.87 &1081.42 &2136.01 &4011.95 &8481.07\\
\hline
\end{tabular}
}
\caption{Memory cost on the ImageNet dataset with different batch sizes.}
\label{tab:memory}
\end{table*}

\begin{table*}[!htpb]
\centering
\begin{tabular}{c|c|c|c|c|c|c|c|c}
\hline
\multirow{2}*{Method} &\multicolumn{7}{c}{Accuracy(\%)/Memory(MB)} \\
\cline{2-9}
& Batch size=1 & Batch size=2 & Batch size=4 & Batch size=8 & Batch size=16 & Batch size=32 & Batch size=64 & Batch size=128 \\
\hline
Vanilla-32bit&22.74/34.64 &85.05/62.68 &88.41/118.77 &90.36/230.94 &91.67/455.29 &92.57/903.99 &92.98/1801.38 &92.84/3596.16\\
EOQ-8bit &89.41/6.39 &92.74/11.64 &93.11/21.64 &93.13/41.63 &92.97/81.63 &92.98/161.61 &94.64/321.59  &93.11/641.53\\
\hline
\end{tabular}
\caption{Results of ResNet-110 on the CIFAR-10 dataset with small batch training.}
\label{tab:small_batch}
\end{table*}

\subsection{Computational Advantages of EOQ}

\begin{figure}[htbp]
    \centering
    \includegraphics[width=0.48\textwidth]{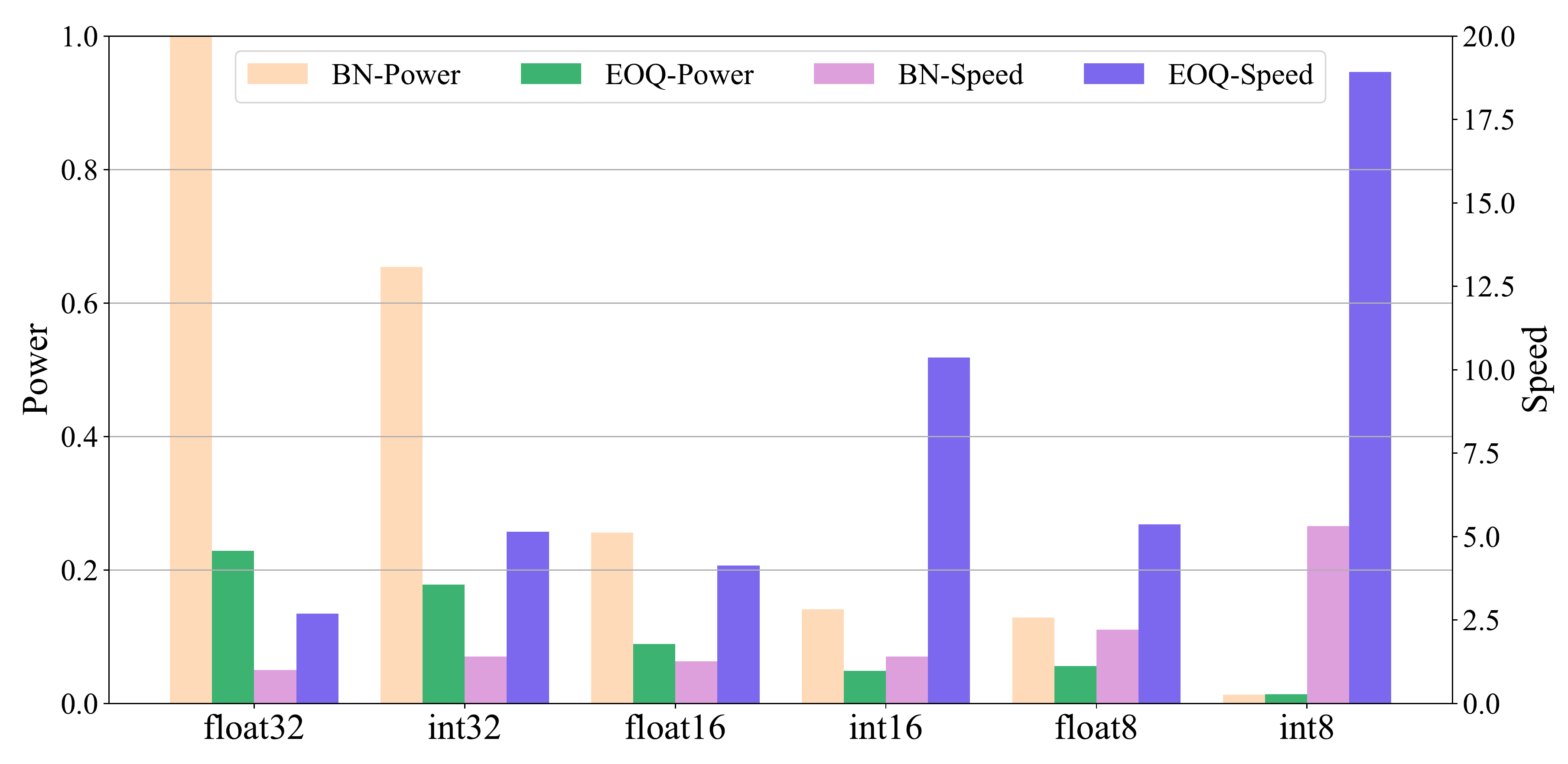}
    \caption{Efficiency comparison of EOQ and BN.}
    \label{fig:eoq_effi}
\end{figure}

\label{sec:advan}

As discussed in previous works, most of the computing costs are concentrated in convolution operations. To measure the hardware overhead more precisely, we use the same FPGA platform (Intel DE5-Net) as WAGEUBN \cite{yang2020training} and L1-BN \cite{wu2018l1} for the test of basic arithmetic operations. Compared with 32-bit floating-point vanilla network, the 8-bit EOQ networks can achieve the same advantages as WAGEUBN \cite{yang2020training}, which is >3$\times$ faster in speed, 10$\times$ lower
in power, 9$\times$ smaller in circuit area in the multiplication operation, and 9$\times$ faster in speed, 30$\times$ lower in power, and 30$\times$ smaller in circuit area in the accumulation operation. What's more, the EOQ takes 2$\times$ smaller weight memory space for that the weights in EOQ stored in memory are with 8-bit while those of WAGEUBN \cite{yang2020training} are with 24-bit.

Another advantage of the EOQ framework lies in the absence of BN. According to L1-BN \cite{wu2018l1}, BN with the high-cost square and root operations in the forward and backward propagation can even reduce the overall computing speed by >47\%, making great challenges for resource-limited ASIC chips and FPGA. Without computing the mean and standard deviation of a mini-batch, EOQ naturally solves the BN trap by a scale factor and bias with much less computation cost and higher hardware parallelism. The efficiency comparison of EOQ and traditional BN is as shown in Figure \ref{fig:eoq_effi}. Only considering the forward propagation, supposing the power and speed of float32 BN are 1, the corresponding data 8-bit EOQ are 0.014 and 18.93. When both BN and EOQ are 8-bit, even though the power of BN and EOQ are similar (0.014 v.s. 0.013), the speed of EOQ is >3$\times$ faster than BN (18.93 v.s. 5.32). As shown in Figure \ref{fig:eoq_effi}, no matter the data type is float32, int32, float16, int16, float8, and int8, EOQ takes great advantages in both power consumption and computing speed. What's more, the memory advantages of EOQ networks over vanilla networks are as shown in Table \ref{tab:memory}. Far beyond this, EOQ only contains low-bit integer multiplication and accumulation operations in the whole training process after removing BN, bringing great convenience for DNN accelerator design.

In summary, combining with the analysis Table \ref{tab:imagenet_acc} and Figure \ref{fig:eoq_effi} comprehensively, we can conclude that the EOQ makes further improvements in network quantization and removing BN, achieving state-of-the-art accuracy with whole 8-bit integer training and great promotion in hardware efficiency.

\subsection{Small Batch Online Training}
\label{sec:small}


For a long time, the large batch size is an important factor for network training. On the one hand, a large batch with enough samples can provide a more accurate gradient for model convergence while the gradient of a small batch may bounce back and forth during training. On the other hand, batch size plays an important role in BN because it is directly related to the computing of the mean and standard deviation of a mini-batch. The large batch size can reduce the impact of extreme samples on the overall results. However, the large batch size is not friendly for online training especially when it comes to portable devices because the limited memory and computational resources can't bear so many input samples in one training iteration. As shown in Table \ref{tab:memory}, though the full 8-bit EOQ can realize >469\% and >37\% less memory consumption compared with the 32-bit and 8-bit vanilla networks, the memory occupied in the training process still raises by >70 times when batch size increases from 1 to 128. At the same time, the small-batch online training has a strong practical significance because it's closer to real applications and can provide resource-limited devices with self-learning ability. In this case, it's essential to study the network training with small batch size.

The EOQ framework has abandoned BN so that it naturally takes advantage in small-batch training. The results of the 32-bit vanilla network using BN and 8-bit EOQ network without BN are as shown in Table \ref{tab:small_batch}. Experiments show that the accuracy of the 8-bit EOQ network is 66.67\%, 7.69\%, and 4.7\% higher than that of the 32-bit floating-point BN network when batch size is 1, 2, and 4, respectively. This fully shows the superiority of EOQ in small-batch training, demonstrating a great perspective in future online training applications of portable devices.

\section{Conclusion}
\label{conclusion}

The two biggest challenging issues named huge computational cost and BN trap have hindered the further development and application of deep learning. To address these two issues, by combining the Fixup initialization technique for removing BN and the quantization technique for reducing computational cost, we propose a general efficient quantization framework named EOQ with specified quantization functions. The EOQ framework has realized a complete 8-bit integer network without BN for the first time. We further give a novel theoretical analysis of the scheme behind removing BN using the Block Dynamical Isometry theory with weaker assumptions. Extensive experiments have been done to verify the effectiveness of EOQ. Results show that the 8-bit EOQ networks can achieve state-of-the-art accuracy and higher hardware efficiency compared with existing works. It also brings great convenience to the deep learning chip design. What's more, we have also borne out that the EOQ framework can be directly used in small batch training, making it closer to the real applications. In summary, the proposed EOQ framework alleviates the problem of huge computational cost and BN trap greatly, demonstrating great perspectives in future online training applications of portable devices.

\section*{ACKNOWLEDGMENT} This work was partially supported by National Key R\&D program (2018AAA0102604, 2018YFE0200200), and key scientific technological innovation research project by Ministry of Education, and Beijing Academy of Artificial Intelligence (BAAI), and a grant from the Institute for Guo Qiang of Tsinghua University, and the Science and Technology Major Project of Guangzhou (202007030006).

\bibliographystyle{unsrt}
\bibliography{ref}

\end{document}